\documentclass[10pt,twocolumn,letterpaper]{article}

\usepackage{iccv}
\usepackage{times}
\usepackage{epsfig}
\usepackage{graphicx}
\usepackage{amsmath}
\usepackage{amssymb}
\usepackage{caption}
\usepackage{subcaption}
\usepackage{threeparttable}
\usepackage{textcomp}
\usepackage{multirow}
\usepackage{booktabs}
\usepackage[pagebackref=true,breaklinks=true,letterpaper=true,colorlinks,bookmarks=false]{hyperref}

\iccvfinalcopy
\def\iccvPaperID{2109}

\ificcvfinal\fi
\begin{document}
\title{Dynamic Multi-path Neural Network}

\author{
    Yingcheng Su\thanks{Equal contribution.}\hspace{4pt}\textsuperscript{1},
    Shunfeng Zhou\footnotemark[1]\hspace{4pt}\textsuperscript{1},
    Yichao Wu\textsuperscript{1},
    Tian Su\textsuperscript{2},
    Ding Liang\textsuperscript{1}\\
    Jiaheng Liu\textsuperscript{3},
    Dixin Zheng\textsuperscript{1},
    Yingxu Wang\textsuperscript{1},
    Junjie Yan\textsuperscript{1},
    Xiaolin Hu\textsuperscript{4}\\
    \textsuperscript{1}SenseTime\hspace{50pt}
    \textsuperscript{2}SEU\hspace{50pt}
    \textsuperscript{3}BUAA\hspace{50pt}
    \textsuperscript{4}THU\\
{
\tt\small suyingcheng@sensetime.com
}
}

\maketitle

\begin{abstract}
Although deeper and larger neural networks have achieved better performance,
the complex network structure and increasing computational cost cannot meet the
demands of many resource-constrained applications.
Existing methods usually choose to execute or skip an entire specific layer,
which can only alter the depth of the network. In this paper,
we propose a novel method called Dynamic Multi-path Neural Network (DMNN),
which provides more path selection choices in terms of network width and
depth during inference. The inference path of the network is determined
by a controller, which takes into account both previous state and object category
information. The proposed method can be easily incorporated into most modern
network architectures. Experimental results on ImageNet and CIFAR-100
demonstrate the superiority of our method on both efficiency and overall
classification accuracy. To be specific, DMNN-101 significantly
outperforms ResNet-101 with an encouraging 45.1\% FLOPs reduction,
and DMNN-50 performs comparably to ResNet-101 while saving 42.1\% parameters.
\end{abstract}

\section{Introduction}

Deep neural networks (DNNs) have dominated the field of computer vision because of superior performance in all kinds of tasks.
It is a tendency that the network architecture is becoming deeper and more complex~\cite{googlenetv1, vggnet, he2016deep, xie2017aggregated, hu2018squeeze} to yield higher accuracy.
However,
the great computing expense of deeper networks contradicts the demands of many resource-constrained applications,
which prefer lightweight networks~\cite{howard2017mobilenets, sandler2018mobilenetv2, zhang2018shufflenet, ma2018shufflenet} to meet limited computation or storage requirement.

\begin{figure}
 \begin{center}
 \begin{subfigure}[b]{0.45\linewidth}
 \centering
  \includegraphics[height=2.0in]{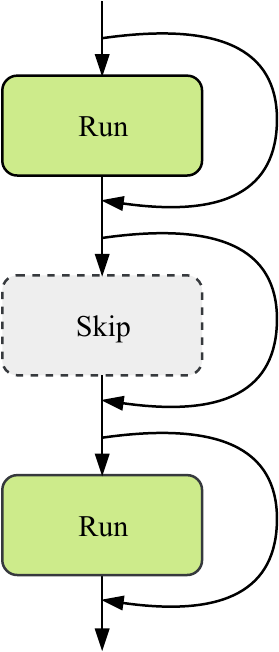}
  \caption{Layer-wise}
  \label{fig:skipnet}
 \end{subfigure}
 \hspace{4pt}
 \begin{subfigure}[b]{0.45\linewidth}
 \centering
  \includegraphics[height=2.0in]{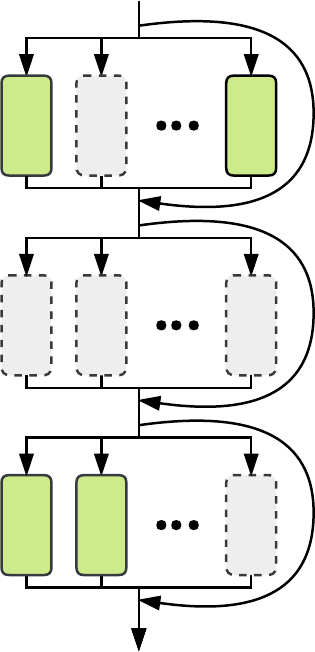}
  \caption{DMNN}
  \label{fig:skipnet_us_4}
 \end{subfigure}
 \end{center}
\vspace{-5pt}
\caption{DMNN provides more flexible and diverse inference paths.}
\vspace{-5pt}
\label{fig:DMNN}
\end{figure}

An elegant solution is to make use of dynamic inference mechanism~\cite{wang2017skipnet, veit2018convolutional, wu2018blockdrop, huang2017multi, dong2017more, figurnov2017spatially, figurnov2016perforatedcnns, teerapittayanon2016branchynet},
reconfiguring the inference path according to the input sample adaptively to meet a better accuracy-efficiency trade-off.
Prevalent dynamic inference techniques are mostly layer-wise methods~\cite{wang2017skipnet, veit2018convolutional, wu2018blockdrop, figurnov2017spatially, teerapittayanon2016branchynet},
as shown in Fig.~\ref{fig:skipnet}.
These methods are usually adopted to determine the execution status of a whole layer at runtime based on a specified mechanism.

All these existing dynamic inference methods only alter the depth of the network. The drawbacks are obvious.
First, 
it is impractical to drop the whole layer/block since some channels of a skipped layer may be useful. 
Second, 
the redundant information between different channels may still exist in the remaining layers. 
A recent study~\cite{zeiler2014visualizing} visualizes the hidden features of CNN models and 
shows the performance contribution from different channels and different layers.
There exists different emphasis on extracting feature among different channels and layers.

In this work, 
we attempt to improve the conventional dynamic inference scheme in terms of both network width and depth
and find an effective forward mechanism for different inputs at runtime from a new perspective of block design. 
We propose Dynamic Multi-path Neural Network (DMNN), 
a novel dynamic inference method that provides various inference path selections.
Fig.~\ref{fig:skipnet_us_4} gives an overview of our approach. 
Different from conventional methods, 
it is expected that each channel has its gate to predict whether to execute or not. 
The primary technical challenge of DMNN is how to design an efficient and effective controller. 


\textbf{Challenge of efficiency.} 
Since DMNN is aimed to conduct channel-wise dynamic evaluation, 
it is ideal for controlling the execution of each channel of the
network at runtime. 
However, this would lead to a significant increase in computational complexity.
Moreover,
as controllers are used at each layer/block of the network, 
they are desirable for lightweight design and generate only a small amount of computational cost.

\textbf{Challenge of effectiveness.} 
The gate control mechanism is similar to SENet~\cite{hu2018squeeze}, 
which adaptively recalibrates channel-wise feature responses by explicitly modeling interdependencies between channels. 
However, 
SENet makes use of soft-weighted sum, 
while DMNN adopts the hard-max mechanism for faster inference while maintaining or boosting accuracy. 
In order to obtain a more reasonable inference path,
it would be better if we take both previous state information and object category into consideration.
Besides,
the resource-constrained loss is also required to make the computational complexity controllable.

To tackle the challenges,
considering that different channels have different representation characteristics, we split the original block of the network
into several sub-blocks. Thus the proposed method provides more optional inference paths. 
A gate controller is introduced to decide whether to execute or skip one sub-block for the current input, 
which only generates minor additional computational cost during inference.
Each block has its controller to control the status of every sub-blocks.
We also carefully design the gate controller to take both previous state information and object category into consideration.
Moreover,
we introduce resource-constrained loss which integrates FLOPs constraint into the optimization process to make the computational complexity controllable.
The proposed DMNN is easy to implement and can be incorporated into most modern network architectures.

The contributions are summarized as follows:
\begin{itemize}
    \vspace{-3pt}
    \item We propose a novel dynamic inference method called \textit{Dynamic Multi-path Neural Network}, which can provide more path selection choices in terms of network width and depth during inference.
    \vspace{-3pt}
    \item We carefully design a gate module controller, which takes into account both previous state and object category information. The resource-constrained loss is also introduced to control the computational complexity of the target network.
    \vspace{-3pt}
    \item Experimental results demonstrate the superiority of our method on both efficiency and overall classification accuracy. To be specific, DMNN-101 significantly outperforms ResNet-101 with an encouraging 45.1\% FLOPs reduction, and DMNN-50 performs comparable results to ResNet-101 with 42.1\% fewer parameters. 
\end{itemize}

\section{Related Work}

\textbf{Adaptive Computation.} 
Adaptive computation aims to reduce overall inference time by changing network topology based on different samples while maintaining or even boosting accuracy. This idea has been adopted in early cascade detectors~\cite{felzenszwalb2010cascade, viola2004robust}, relying on extra prediction modules or handcrafted control strategies. Learning based layer-wise dynamic inference schemes are widely investigated in the field of computer vision. Early prediction models like BranchyNet~\cite{teerapittayanon2016branchynet}and Adaptive Computation Time~\cite{figurnov2017spatially} adopt branches or halt units to decide whether the model could stop early. Some works use gate mechanism to determine the execution of a specific block. Wang et al.~\cite{wang2017skipnet} propose SkipNet which uses a gating network to selectively skip convolutional blocks based on the activations of the previous layer. A hybrid learning algorithm that combines supervised learning and reinforcement learning is used to address the challenges of non-differentiable skipping decisions. Wu et al.~\cite{wu2018blockdrop} propose BlockDrop and also make use of a reinforcement learning setting for the reward of utilizing a minimal number of blocks while preserving recognition accuracy. ConvNet-AIG is proposed in~\cite{veit2018convolutional}, which utilizes the Gumbel-Max trick~\cite{gumbel1954statistical} to optimize the gate module. However, the block-wise method can only alter the depth of the network, which could be too rough as some channels of an abandoned block may be useful. 

On the other hand, the channel-wise method can manually adjust the number of active channels of a specific model. However, as far as we know, only~\cite{yu2018slimmable} is similar to such a method. The proposed Slimmable Neural Networks can adjust its width on the fly according to the on-device benchmarks and resource constraints. Strictly speaking, it is not a dynamic process as the procedure of choosing the active channels is finished before inference. Moreover, the pre-defined width multipliers negatively affect the flexibility of the dynamic inference mechanism. Our work is close to~\cite{veit2018convolutional}. However, we attempt to combine the merits of both the above two methods and propose a novel dynamic inference method which can provide more path selection choices in terms of network width and depth.

\textbf{Model Compression.} The great computing expense of deeper networks contradicts the demands of many resource-constrained applications, such as mobile platforms, therefore, reducing storage and inference time also plays an important role in deploying top-performing deep
neural networks. Lots of techniques are proposed to attack this problem, such as pruning~\cite{he2017channel, molchanov2016pruning, wen2016learning}, distillation~\cite{hinton2015distilling, polino2018model}, quantization~\cite{han2015deep, wei2018quantization, jacob2018quantization}, low-rank factorization~\cite{ioannou2015training}, compression with structured matrices~\cite{cheng2015exploration} and network binarization~\cite{courbariaux2016binarized}. However, these works are usually applied after training the initial networks and generally used as post-processing, while DMNN could be trained end-to-end without well-designed training rules.

On the other hand, lightweight architectures play important roles in various real scenarios, such as MobileNet~\cite{howard2017mobilenets, sandler2018mobilenetv2} and ShuffleNet~\cite{zhang2018shufflenet, ma2018shufflenet}. In this paper, by applying our methods, we prove even compact model like MobileNetV2 could be further improved. 


\section{Methodology}

In this section,
we introduce the proposed dynamic multi-path neural network (DMNN) in detail,
including the subdivision of the block,
the architecture of the controller and the optimization approach.

\subsection{Block Subdivision}
It is ideal for controlling the execution of each channel of the network at runtime.
However,
this would lead to a significant increase in computational complexity.
In this work,
we divide the origin block of the network into several sub-blocks,
and each sub-block has its switch to decide whether to execute or not,
resulting to a dynamic inference path for different samples.
We interpret optimizing the network structure as executing or skipping of each sub-block during the inference stage.

\begin{figure}
  \begin{center}
    \centering
    \begin{subfigure}[b]{0.8\linewidth}
        \centering
        \includegraphics[width=\linewidth]{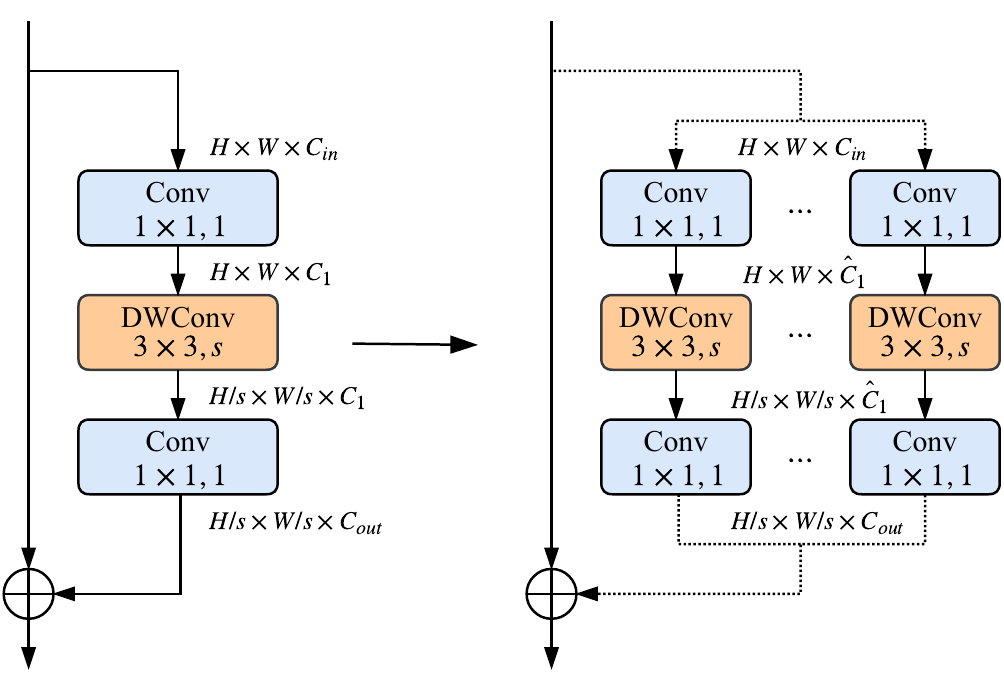}
        \vspace{-20pt}
        \caption{MobileNetV2 block}
        \label{fig:orig_mobile}
    \end{subfigure}
    \\
    \begin{subfigure}[b]{0.8\linewidth}
        \centering
        \includegraphics[width=\linewidth]{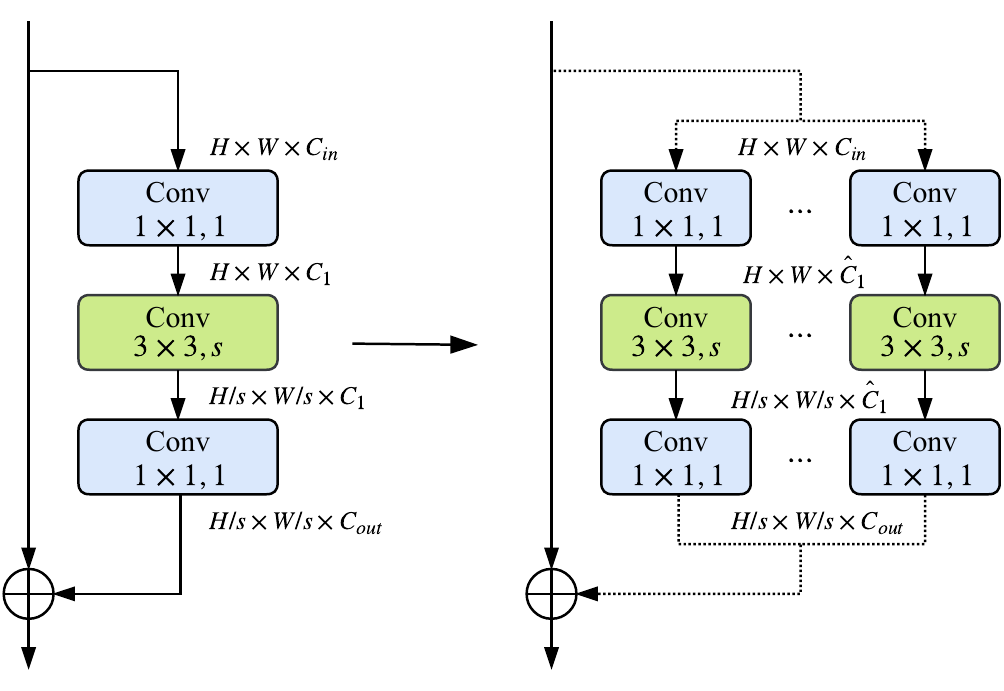}
        \vspace{-20pt}
        \caption{ResNet block}
        \label{fig:orig_res}
    \end{subfigure}
  \end{center}
  \vspace{-20pt}
  \caption{Subdivision strategy.}
  \vspace{-10pt}
  \label{fig::subdivide}
\end{figure}

A key issue is how to split one block into $N$ sub-blocks.
The guiding principle is that the parameters of the new block must be consistent with or approximate to the original block for fair comparison.
Fig.~\ref{fig::subdivide} shows the subdivision of blocks of MobileNetV2 and ResNet.

For the block of MobileNetV2,
we divide the origin block into $N$ sub-blocks,
the expansion ratio of each sub-block is set to $E/N$.
Thus the sum of every sub-block's computation and parameters are the same with the original block since it only consists of pixel-wise convs and depth-wise convs,
more detail can be seen in Fig~\ref{fig:orig_mobile}.

While for ResNet,
it is not that straightforward.
As shown in Fig.~\ref{fig:orig_res},
suppose the shape of the input tensor is $H \times W \times C_{in}$,
the output channels of each conv operation are $C_1, C_2, C_{out}$.
The parameter of the original block is
\begin{equation}
    C_{in}\cdot C_1 + 9\cdot C_1 \cdot C_2 + C_2 \cdot C_{out}.
    \label{eq:old_param}
\end{equation}
The original block is then split into $N$ sub-blocks.
The output channels of each sub-block are $\hat{C_1}, \hat{C_2}, C_{out}$.
Then the parameter becomes
\begin{equation}
    N \cdot \left(C_{in}\cdot \hat{C_1} + 9 \cdot \hat{C_1} \cdot \hat{C_2}+ \hat{C_2} \cdot C_{out} \right).
    \label{eq:new_param}
\end{equation}
If we simply set the number of channels of each sub-block to $1/N$ of the origin blocks, i.e.
$\hat{C_1} = C_1 / N, \hat{C_2} = C_2 / N$, Eqn.~\ref{eq:new_param} can be rewritten as follow:
\begin{equation}
     C_{in}\cdot C_1 + 9\cdot C_1 \cdot C_2 / N + C_2 \cdot C_{out},
\end{equation}
which is not equal to Eqn.~\ref{eq:old_param}.
Thus,
to make subsequent extensive studies fair,
we make minor modifications to ResNet
and design the corresponding DMNN version to make Eqn.~\ref{eq:old_param} $\approx$ Eqn.~\ref{eq:new_param}.
For saving space,
the detailed architecture of DMNN-50 can be referred to supplementary materials.

\subsection{The Architecture of Controller}
\label{section:CAM}

\begin{figure}[t]
\begin{center}
    \begin{subfigure}[b]{0.8\linewidth}
        \includegraphics[width=\linewidth]{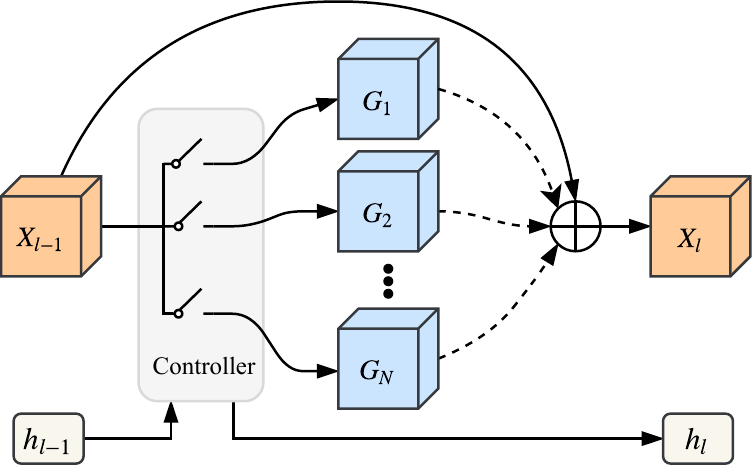}
        \vspace{-18pt}
        \caption{The overview of our gate mechanism.}
        \label{fig:controller_overview}
    \end{subfigure}
    \begin{subfigure}[b]{0.8\linewidth}
        \includegraphics[width=\linewidth]{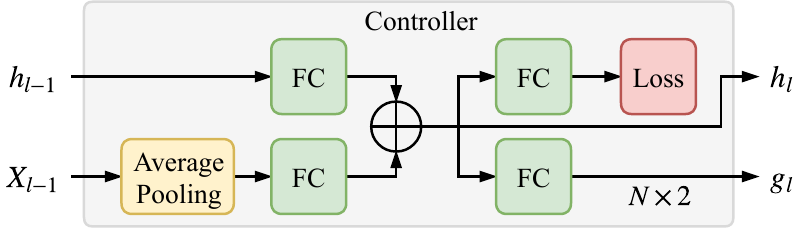}
        \vspace{-18pt}
        \caption{The architecture of controller.}
        \label{fig:controller_module}
    \end{subfigure}
\end{center}
\vspace{-20pt}
\caption{The framework of dynamic paths selection.}
\vspace{-10pt}
\label{fig:controller_architecture}
\end{figure}

The controller is elaborately designed to predict the status of each sub-block (on/off) with an minimal cost.
It is the inference paths optimizer of DMNN.
An overview of the dynamic path selection framework is shown in Fig.~\ref{fig:controller_architecture}.
Given an input image,
its forward path is determined by the gate controllers
and Fig.~\ref{fig:controller_overview} shows the gate mechanism of DMNN. 
Suppose we split $l$-th block into $N$ sub-blocks,
the output of $l$-th block is the combination of the outputs of an identity connection and $N$ sub-blocks.
Formally,
\begin{equation}
    \boldsymbol{X_l}=\boldsymbol{X_{l-1}} + \sum_{n}^{N}s_{l,n}\mathcal{F}_{l,n}(\boldsymbol{X_{l-1}}),
\end{equation}
where $\boldsymbol{X_l}$ is output of  $l$-th block, $s_{l,n}\in\{0,1\}$ refers to the off/on status
which is predicted by the controller.
$\mathcal{F}_{l,n}(\boldsymbol{X_{l-1}})$ refers to the output of $n$-th sub-block of $l$-th block.

\textbf{Spatial and previous state information embedding}.
On the one hand,
the control modules make decisions based on the global spatial information,
and we achieve this process by applying global average pooling to compress the high dimension features to one dimension along channels.
We further use a fully connected layer followed by an activation layer to map the pooling features to low-dimensional space.
Specifically,
$\boldsymbol{X_{l-1}} \in \mathbb{R}^{H \times W \times C}$ represents the input features of $l$-th block,
we calculate the $c$-th channel statistic by
\begin{equation}
    z_c = \frac{1}{H \cdot W} \sum_{i=1}^{H} \sum_{j=1}^{W}x_{i,j,c}^{l-1},
\end{equation}
The final embedding feature $\boldsymbol{V_{l-1}} \in \mathbb{R}^d$ is
\begin{equation}
    \boldsymbol{V_{l-1}} = \mathcal{F}(\boldsymbol{z}, \boldsymbol{W_1}) = \sigma(\boldsymbol{W_1}\boldsymbol{z}),
\end{equation}
where $\boldsymbol{z}=\left[z_1, z_2,\cdots, z_c \right]$, $\boldsymbol{W_1} \in \mathbb{R}^{d \times c}$,
$\sigma$ is the ReLU~\cite{glorot2011deep} function,
$d$ is the dimension of the hidden layer.

On the other hand,
there are some connections between the current controller and the previous controllers.
Thus the integration of previous state information is also crucial.
We first employ a fully connected layer followed by ReLU function to map the previous state hidden features into the same subspace with $\boldsymbol{V_{l-1}}$.
Then we perform an addition operation on the hidden feature and $\boldsymbol{V_{l-1}}$ to get the result of the current state.
Formally, 
\begin{equation}
\begin{split}
    \boldsymbol{h_{l-1}^{'}} &= \mathcal{F}(\boldsymbol{h_{l-1}},\boldsymbol{W_2}) = \sigma(\boldsymbol{W_2}*\boldsymbol{h_{l-1}}), \\
    \boldsymbol{h_l} &= \boldsymbol{V_{l-1}} + \boldsymbol{h_{l-1}^{'}},
\end{split}
\end{equation}
where $\boldsymbol{W_2} \in \mathbb{R}^{d \times d}$,
$\sigma$ represents the ReLU function.
Bias terms are omitted for simplicity.
The status predictions of each sub-block at $l$-th block are made through $\boldsymbol{h_l}$ by using a softmax trick which we will introduce in section \ref{section:optimization}.

\textbf{Softmax Trick with Gumbel Noise.}
To decide whether to execute or omit a sub-block is inherently discrete and therefore non-differentiable.
In this work,
we use softmax trick with gumbel noise to solve this problem,
which has been proved to be successful in~\cite{veit2018convolutional}.
Formally,
let $N$ be the number of sub-blocks and $\boldsymbol{g_l}=\boldsymbol{W_3h_l}+\boldsymbol{b_3}$, $\boldsymbol{W_3} \in \mathbb{R}^{2N \times d}$, $\boldsymbol{b_3}$ is the bias term.
$\boldsymbol{g_l}$ is then reshaped to $N\times 2$ for the final predictions.
The activation can be written as follows
\begin{equation}
    \boldsymbol{s_l} = \arg \max \left( \text{softmax} \left(\boldsymbol{g_l} + \Delta \right) \right),
    \label{eq:decision}
\end{equation}
where $\boldsymbol{s_l}=\left[s_{l,1}, s_{l,2}, \cdots, s_{l,N} \right]$ refers to the status of each sub-block of $l$-th block,
and $\Delta \sim \text{ Gumbel}(0, 1)$ is a random noise following the Gumbel distribution,
which can increase the stability of the training process of our network.

\textbf{Supervised learning of controller}.
Deep CNNs compute feature hierarchies in each layer
and produce feature maps with different depths and resolutions.
This can also be considered as a feature extraction process from coarse to fine.
The proposed DMNN has a diversity of inference paths,
and we hope that different classes would select different paths.
However,
if the path selection mechanism is trained only by optimizing the classification loss at the last layer,
it will be difficult for the controller to learn the category information.
To solve this problem,
we introduce category loss to each controller to enable all of them to become category-aware.
Considering that predicting each class as a different category by the controller is computationally expensive,
we cluster samples into fewer categories than original classes.
For the ImageNet dataset~\cite{imagenet_cvpr09},
we cluster all the 1000 classes samples into 58 big categories with the help of the hierarchical structure of ImageNet provided in ~\cite{imagenet_cvpr09}.
For the CIFAR-100 dataset~\cite{krizhevsky2009learning},
it groups the 100 classes into 20 superclasses.
We use the 20 superclasses as the big categories directly.
Then cross entropy loss is employed to supervise all controllers as shown in Fig.~\ref{fig:controller_module}.
Formally,
the category loss of $l$-th controller can be written as follow
\begin{equation}
    \mathcal{L}_{l}= \sum_{j}^{K}k_j \log(p_j),
\end{equation}
where $p_j$ represents the probability of $j$-th class.
$k_j=1$ if $j$ is the ground-truth class and 0 otherwise,
$K$ indicates the number of categories.
It is worth noting that the loss weights of each block's controller are not always equal since the features of different layers have different semantic information.
Deep layers have a stronger semantic information than shallow layers.
In DMNN-50,
there are four stages composed of 3, 4, 6, 3 stacked blocks respectively,
resulting in 16 controllers.
The loss weight of the first stage is set to 0.0001,
and it will increase by a factor of 10 in the next stages.
DMNN-101 follows the same principle.
The loss of supervised controller can be represented as follows
\begin{equation}
    \mathcal{L}_{ctg} = \sum_l^L \alpha_l \mathcal{L}_{l},
\end{equation}
where $\alpha_l$ denotes the loss weight of $l$-th controller and $L$ denotes the number of blocks.
The category information will be removed after training,
so it will not generate any extra computational burden during testing.

The controller is desirable for its lightweight characteristic during the optimization of network structure.
The dimension of the hidden layer $d$ is set to 32 in all experiments.
This setting generates only little computational cost and can be omitted compared to the whole computation of the network.
If we take DMNN-50 as an example,
the total 16 controllers only generate about 0.02\% FLOPs of the original ResNet-50.

\subsection{Optimization}
\label{section:optimization}
\textbf{Resource-constrained Loss.} The resource constraint comes from two aspects: the block execution rate and the total FLOPs. The execution rate of each block in a mini-batch is used to constrain the average block activation rate to the target rate $e$. Let $z_l$ denotes the execution rate of $l$-th block within a mini-batch, we define the execution rate $z_l$ as
\begin{equation}
    z_l = \frac{\sum_i^N b_i} {B \cdot N},
\end{equation}
where $B$ is the mini-batch size, $b_i$ is the executed number of $i$-th sub-block within a mini-batch. The total execution rate loss can be written as follow
\begin{equation}
\mathcal L_{exec} = \sum_l^L \left(e-z_l\right)^2.
\end{equation}
The other constraint is the total FLOPs. To meet the desired FLOPs, we explicitly introduce the target FLOPs rate to the loss function. In each mini-batch, we compute the actual FLOPs via
\begin{equation}
    f=\sum_{l}^{L}\sum_{i}^{N} \frac{b_i}{B} \cdot f_{l,i},
\end{equation}
where $f_{l,i}$ indicates the FLOPs of $i$-th sub-block at $l$-th block of the network. The FLOPs loss can be formulated as 
\begin{equation}
  \mathcal{L}_{flops} = \left (\frac {f} {f_{total}} - r \right )^2,
\end{equation}
where $f_{total}$ and $f$ represent the full FLOPs and the actual execution FLOPs of the network respectively, and $r$ denotes the target FLOPs rate. We set $e=r$ in all experiments since they have strong positive correlation and similar values. Thus, the resource-constrained loss is defined as
\begin{equation}
    \mathcal{L}_{res} = \mathcal{L}_{exec} + \mathcal{L}_{flops}.
\end{equation}
The total training loss is
\begin{equation}
\mathcal{L}_{total} = \alpha_1 \mathcal{L}_{ctg}
+ \alpha_2 \mathcal{L}_{res}
+ \alpha_3 {\mathcal{L}_{cls}},
\end{equation}
where $\mathcal{L}_{cls}$ is the classification loss. In our experiments $\alpha_1=\alpha_2=\alpha_3=1$. The joint loss would be optimized by mini-batch stochastic gradient descent.

\begin{table*}[t]
\begin{center}
\begin{threeparttable}
\begin{tabular}{lcccc}
\toprule[1pt]
Model & Top-1 Err. (\%) 
&  Params ($10^6$)  &  FLOPs ($10^9$)   & FLOPs Ratio (\%)\\
\hline
ResNet-50~\cite{he2016deep}
                      & 24.7  
                      & 25.56 & 3.8  & -     \\
ResNet-50 (PyTorch Official)~\cite{pytorch_models}
                      & 23.85  
                      & 25.56 & 3.96  & 100.0     \\
ResNet-50\tnote{\dag} (ours)
                      & 23.51 
                      & 25.56 & 3.96 & 100.0 \\
ResNet-50 + Pruning~\cite{molchanov2016pruning}
                      & 23.91 
                      & 20.45 & 2.66 & 70.0 \\
ResNeXt-50 [$2\times40d$] ~\cite{xie2017aggregated} 
                      & 23.0
                      & 25.4 & 4.16 & 105.1 \\
ResNeXt-50 [$4\times24d$] ~\cite{xie2017aggregated}
                      & 22.6
                      & 25.3 & 4.20 & 106.1 \\
ConvNet-AIG-50 [$t=0.7$] ~\cite{veit2018convolutional}
                      & 23.82 
                      & 26.56 & 3.06 & 77.3 \\
S-ResNet-50-0.75~\cite{yu2018slimmable}
                      & 25.1  
                      & 19.2  & 2.3  & 58.1 \\
DMNN-50, $N=2$ [$r=0.4$]   & 24.06 
                      & 24.67 & 2.07 & 52.3 \\
DMNN-50, $N=2$ [$r=0.5$]   & 23.50 
                      & 24.67 & 2.28 & 57.6 \\
DMNN-50, $N=2$ [$r=0.6$]   & 23.22 
                      & 24.67 & 2.52 & 63.6 \\
DMNN-50, $N=2$ [$r=0.7$]   & 22.57 
                      & 24.67 & 3.12 & 78.8 \\
DMNN-50, $N=3$ [$r=0.7$]   & 22.54 
                      & 25.81 & 3.16 & 79.8 \\
DMNN-50, $N=4$ [$r=0.7$]   & \textbf{22.32} 
                      & 25.81 & 3.17 & 80.1 \\
\hline
ResNet-101~\cite{he2016deep}
                      & 23.6  
                      & 44.54 & 7.6  & - \\
ResNet-101 (PyTorch Official)~\cite{pytorch_models}
                      & 23.63  
                      & 44.55 & 7.67  & 100.0 \\
ResNet-101\tnote{\dag} (ours)
                      & 22.02 
                      & 44.55 & 7.67 & 100.0 \\
ResNeXt-101 [$2\times40d$] ~\cite{xie2017aggregated}
                      & 21.7 
                      & 44.46 & 7.9 & 103.0 \\
ConvNet-AIG-101[$t=0.5$]~\cite{veit2018convolutional}
                      & 22.63 
                      & 46.23 & 5.11 & 66.6 \\
DMNN-101, $N=2$ [$r=0.3$]  & 22.82 
                      & 43.12 & 2.48 & 32.3 \\
DMNN-101, $N=2$ [$r=0.5$]  & 21.95 
                      & 43.12 & 4.21 & 54.9 \\
DMNN-101, $N=2$ [$r=0.7$]  & \textbf{21.43} 
                      & 43.12 & 5.57 & 72.6 \\
\bottomrule[1pt]
\end{tabular}
\begin{tablenotes}
\footnotesize
\item[\dag] Our implementations of ResNet-50, ResNet-101, DMNN-50, DMNN-101 use $stride=2$ in conv$3\times3$ layers just as the PyTorch community does~\cite{pytorch_models} which is slightly different from the original paper.
\end{tablenotes}
\end{threeparttable}
\end{center}
\vspace{-10pt}
\caption{Comparison on heavyweight networks on ImageNet. We compare our DMNNs with the heavyweight networks ResNet-50, ResNet-101, and other dynamic networks ConNet-AIGs and slimmable network. Results show that our models outperform the other models in both accuracy and computational complexity.}
\vspace{-5pt}
\label{tab:resnet result}
\end{table*}

\section{Experiments}

\begin{figure}[t]
\begin{center}
\includegraphics[width=0.7\linewidth]{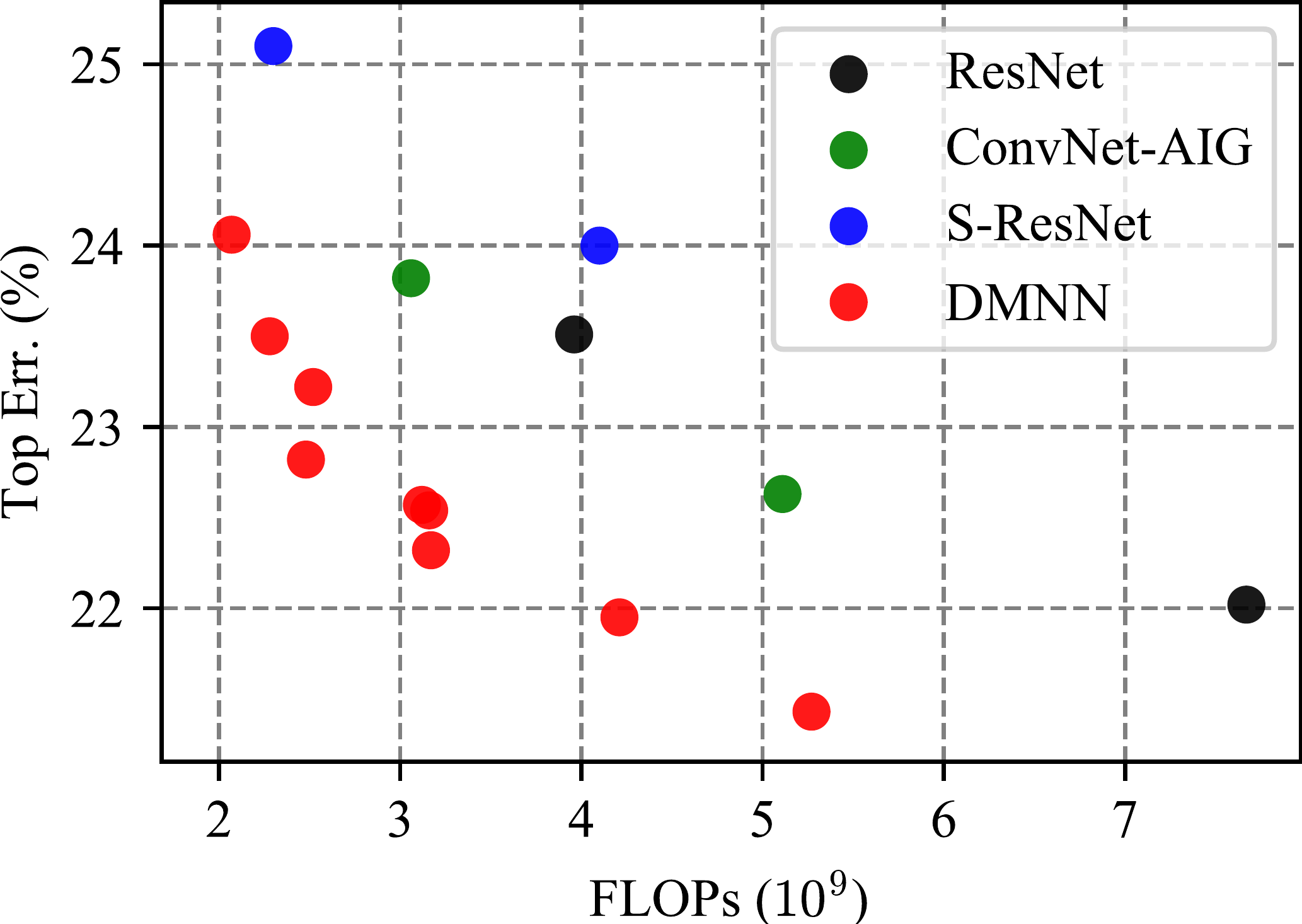}
\end{center}
\vspace{-10pt}
   \caption{
      {Top-1 error vs. FLOPs on ImageNet.} The proposed DMNN models outperform other methods by a large margin in both computational cost and accuracy.
   }
\vspace{-10pt}
\label{fig:flops}
\end{figure}

In this section, we evaluate the performance of the proposed DMNN on benchmark datasets including ImageNet and CIFAR-100. 


\subsection{Training Setup}

\textbf{ImageNet.} The ImageNet dataset~\cite{imagenet_cvpr09} consists of 1.2 million training images and 50K validation images of 1000 classes. We train networks on the training set and report the top-1 errors on the validation set. We apply standard practice and perform data augmentation with random horizontal flipping and random-size cropping to 224$\times$224 pixels. We follow the standard Nesterov SGD optimizer with momentum 0.9 and a mini-batch of 256. The cosine learning rate scheduler is employed for better convergence and the initial learning rate is set to 0.1. For different scale models, We use different weight decays, 0.0001 for ResNet and 0.00004 for MobileNet.  All models are trained for 120 epochs from scratch.

\textbf{CIFAR-100.} The CIFAR-100 datasets~\cite{krizhevsky2009learning} consist of 60,000 color images of 10, 000 classes. They are split into the training set and testing set by the ratio of 5:1. Considering the small size of images ($32 \times 32$) in CIFAR, we follow the same setting as \cite{he2016deep} to construct our DMNNs for a fair comparison. We augment the input image by padding 4 pixels on each side with the value of 0, followed by random cropping with a size of $32 \times 32$ and random horizontal flipping. We train the network using SGD with the momentum of 0.9 and weight decay of 0.0001. The mini-batch size is set to 256, and the initial learning rate is set to 0.1. We train the networks for 200 epochs and divide the learning rate by 10 twice, at the 100th epoch and 150th epoch respectively.

\subsection{Performance Analysis}
We compare our method with ResNet~\cite{he2016deep}, ResNeXt~\cite{xie2017aggregated},
MobileNetV2~\cite{sandler2018mobilenetv2}, pruning method~\cite{molchanov2016pruning} and other dynamic 
inference methods~\cite{yu2018slimmable, veit2018convolutional}. 
We denote $N$ as the number of sub-blocks of each block, $r$ as the FLOPs target rate.

\begin{table*}[t]
\begin{center}
 \begin{threeparttable}
\begin{tabular}{lcccc}
\toprule[1pt]
Model & Top-1 Err. (\%) 
&  Params ($10^6$)  &  FLOPs ($10^9$)   & FLOPs Ratio (\%)\\
\hline
MobileNet V2~\cite{sandler2018mobilenetv2} 
                                & 28.0 
                                & 3.47  & - & -     \\
MobileNet V2 (ours)
                                & 28.09
                                & 3.50\tnote{\dag} & 0.30 & 100.0 \\
S-MobileNet V2-0.75~\cite{yu2018slimmable} 
                                & 31.1 
                                & 2.7  & 0.23 & 76.7 \\
DMNN-MobileNetV2, $N=2$ [$r=0.7$]   & 28.30 
                               & 3.63 & 0.22 & 73.3 \\
DMNN-MobileNetV2, $N=2$ [$r=0.8$]   & 28.15 
                               & 3.63 & 0.24 & 80.0 \\
DMNN-MobileNetV2, $N=2$ [$r=0.9$]   & \textbf{27.74}
                               & 3.63 & 0.27 & 90.0 \\
\hline
MobileNetV2 (1.4)~\cite{sandler2018mobilenetv2}
                                    & 25.3 
                                    & 6.06  & - & - \\
MobileNetV2 (1.4) (ours)
                                    & 25.30
                                    & 6.09\tnote{\dag} & 0.57 & 100.0 \\
DMNN-MobileNetV2 (1.4), $N=2$ [$r=0.7$] & 26.03
                                   & 6.29 & 0.42 & 73.7 \\
DMNN-MobileNetV2 (1.4), $N=2$ [$r=0.8$] & 25.53
                                   & 6.29 & 0.47 & 82.5 \\
DMNN-MobileNetV2 (1.4), $N=2$ [$r=0.9$] & \textbf{25.26}
                                   & 6.29 & 0.52 & 91.2 \\
\bottomrule[1pt]
\end{tabular}
\begin{tablenotes}
\footnotesize
\item[\dag]
Our implementation of MobileNet V2 is based on PyTorch and its parameter quantities are counted by PyTorch Summary~\cite{pytorch_summary}.
\end{tablenotes}
\end{threeparttable}
\end{center}
\vspace{-10pt}
\vspace{-2pt}
\caption{Comparison on lightweight networks on ImageNet. Our DMNNs based on MobileNetV2 can achieve remarkable results comparing to other lightweight models.}
\vspace{-10pt}
\label{tab:mobilenet result}
\end{table*}

\textbf{Performance on heavy networks.} 
Tab.~\ref{tab:resnet result} shows that our DMNN achieves remarkable results compared to other heavy models on ImageNet.
First of all, we compare DMNN with ResNet. 
When $N=2$, $r=0.5$, our DMNN-50 achieves similar performance with ResNet-50 but saves more than \textbf{42.4}\% FLOPs.
When we set $N=4$, $r=0.7$, our DMNN-50 further reduces 1.19\% Top-1 error while still saving \textbf{19.9}\% FLOPs.
Our DMNN-101 outperforms ResNet-101 and save \textbf{45.1}\% FLOPs in the same time when we set $N=2$, $r=0.5$.
The above comparison demonstrates that DMNN can greatly reduce FLOPs and improve the accuracy when compared to the models with similar parameters.  
On the other hand, 
DMNN-50 achieves even better performance than origin ResNet-101 (closely to ResNet-101 by our implementation), with \textbf{42.0}\%
parameter size reduction, 
which indicates that DMNN can greatly save the parameters and is feasible for pratical model deployment. 

Then, 
we make comparason with DMNN and stronger baseline models ResNeXt.
As we set $r=0.7$, 
our method is superior to ResNeXt-50 ($2\times40d$) in both FLOPs and accuracy.
When we set $N=4$, $r=0.7$, 
our DMNN-50 reduces 0.28\% Top-1 error while still saves \textbf{24.5}\% FLOPs.
Our DMNN-101 outperforms ResNet-101 and save \textbf{45.1}\% FLOPs in the same time when we set $N=2$, $r=0.5$.
Similar result can be found while comparing to ResNeXt-101 ($2\times40d$) if we set $r=0.7$.
DMNN shows great superiority over ResNeXt mainly because of better control for different convolution groups. 

Our method outperforms ConvNet-AIG~\cite{veit2018convolutional} in both accuracy and computational complexity, 
demonstrating that multi-path design is more elaborate and superior than roughly skipping the whole block.
Especially, our DMNN-50 with $N=2$ and $r=0.4$ achieves comparable performance with ConvNet-AIG yet greatly
reduces the FLOPs by approximately \textbf{33.3}\%.
Fig~\ref{fig:flops} shows the trade-off between computational cost and accuracy of our DMNN while
comparing to other dynamic inference methods including slimmable neural network S-ResNet~\cite{yu2018slimmable}.
Meanwhile, 
as an end-to-end method, DMNN shows great advantages over post process method such as pruning methods.

We also conduct experiments on CIFAR-100 dataset, 
as shown in Tab.~\ref{tab:cifar}. It can be seen that DMNN-50
with $N=4$ and $t=0.7$ can even outperform ResNet-50 by 1.4\% on 
CIFAR-100 with only \textbf{78.7}\% FLOPs.

\textbf{Performance on lightweight networks.} 
We apply DMNN to lightweight network MobileNetV2, as shown in Tab.~\ref{tab:mobilenet result}. 
Although similar conclusions can be obtained, 
it is normal that the improvements is not as large as heavy models because of the compact structures. 
Specially, our DMNN with $N=2$ and $r=0.9$ can save 10.0\% FLOPs and achieves better top1 error than MobileNetV2. The proposed method is also better than other dynamic inference methods.


In summary, our method performs superbly in accuracy and computational complexity for both heavy and lightweight networks, which demonstrates its great applicability to different networks and robustness on different datasets.

\begin{table}[t]
\addtolength{\tabcolsep}{-4.2pt}
\begin{center}
\begin{tabular}{lcc}
    \toprule[1pt]
    Model & FLOPs ($10^9$) & Top-1 Err. (\%) \\
    \hline
    ResNet-50~\cite{he2016deep} & 0.33 & 27.55 \\
    \hline
    DMNN-50, $N=2$ [$t=0.5$] & 0.18 & 28.24 \\
    DMNN-50, $N=2$ [$t=0.7$] & 0.22 & 27.34 \\
    DMNN-50, $N=4$ [$t=0.7$] & 0.26 & \textbf{26.15} \\
    \bottomrule[1pt]
\end{tabular}
\end{center}
\vspace{-10pt}
\vspace{-2pt}
\caption{Test error on CIFAR-100. The DMNNs reduce 1.4\% Top-1 error while saving about 21.2\% FLOPs.}
\vspace{-10pt}
\label{tab:cifar}
\end{table}

\begin{table}
\begin{center}
    \begin{tabular}{l|cc|cc}
    \toprule[1pt]
    Method & PREV & CAT & Top-1 Err. (\%) \\
    \hline
    ResNet-50~\cite{he2016deep} & & & 23.51 \\
    \hline
    DMNN-50 & & & 23.25 \\
    DMNN-50 & $\surd$ & & 23.09 \\
    DMNN-50 & & $\surd$ & 23.20 \\
    DMNN-50 & $\surd$ & $\surd$ & \textbf{22.57}\\
    \bottomrule[1pt]
    \end{tabular}
\end{center}
\vspace{-10pt}
\vspace{-2pt}
\caption{The effectiveness of well-designed controller with supervised learning on ImageNet.
The FLOPs target is set to 0.7, and the number of sub-block $N$ is set to 2.
``PREV'' represents employing previous state features
and ``CAT'' represents employing supervised learning in this table.}
\vspace{-10pt}
\label{tab:cacs}
\end{table}

\subsection{Ablation Study}
\textbf{Effectiveness of the gate controller.}
In order to show the effectiveness of the controllers,
we conduct four groups of experiments on ImageNet dataset with different configurations.
Tab.~\ref{tab:cacs} shows the comparison of different models. 
If we employ previous features and
supervised learning separately, additional promotions are obtained. After aggregating
these two improvements, we can boost the performance by 0.68\%, demonstrating the benefits of previous state information embedding
and supervised learning of controllers. 
It is worth noting that it only introduces a fully connected layer
with 32 hidden neurons while applying previous controller's features, the additional computation
cost can be omitted. The supervised learning of the controllers may generate minor additional
computational cost during training, yet it will be removed at the testing stage.

\begin{figure}[t]
    \begin{center}
    \includegraphics[width=0.7\linewidth]{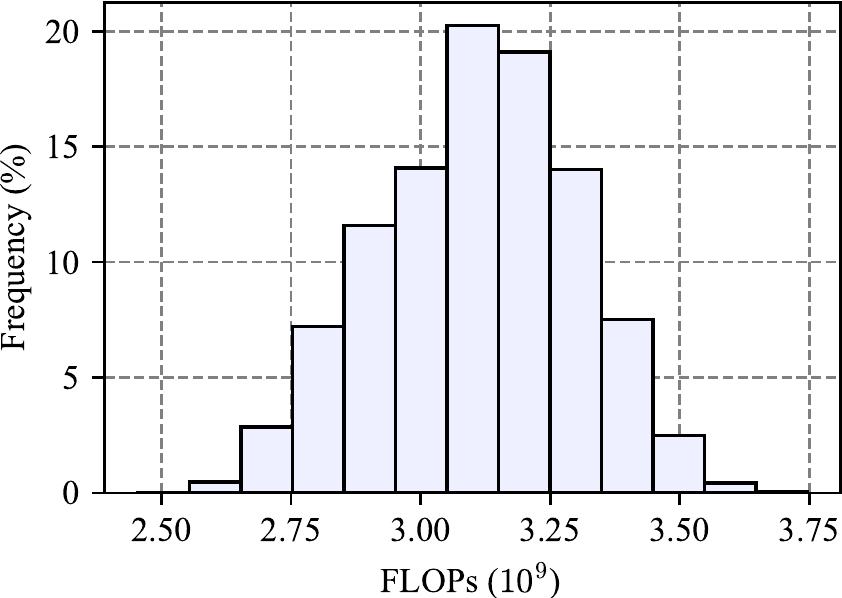}
    \end{center}
    \vspace{-10pt}
    \vspace{-3pt}
    \caption{
      {Distribution of FLOPs on the ImageNet validation set using DMNN-50 with $N=2$, $r=0.7$.}
    }
    \vspace{-10pt}
    \label{fig:dist}
\end{figure}

\begin{figure*}
    \centering
    \includegraphics[width=\linewidth]{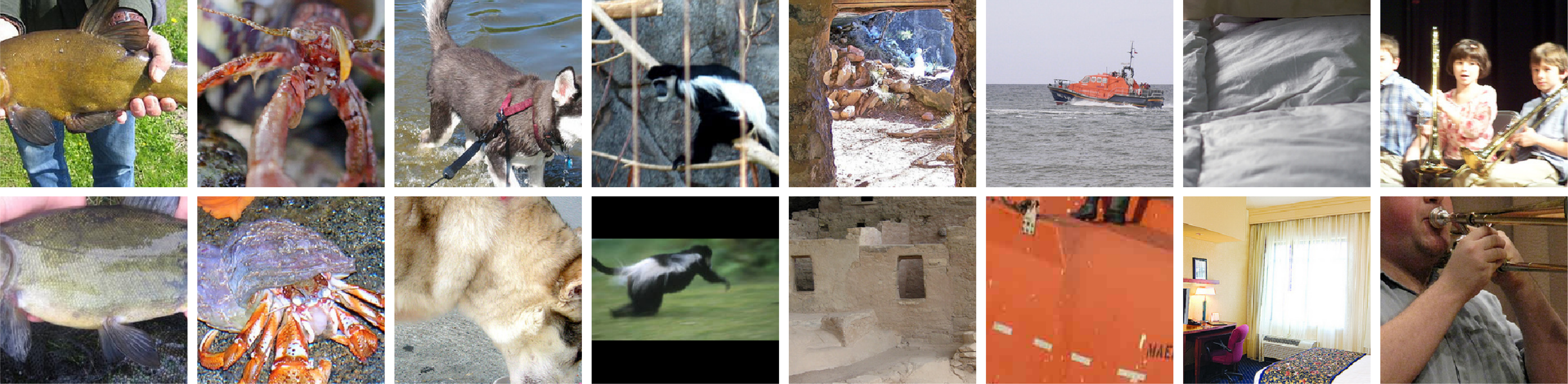}
    \begin{subfigure}[b]{0.120\linewidth}
        \caption{tench}
    \end{subfigure}
    \begin{subfigure}[b]{0.120\linewidth}
        \caption{hermit crab}
    \end{subfigure}
    \begin{subfigure}[b]{0.120\linewidth}
        \caption{malamute}
    \end{subfigure}
    \begin{subfigure}[b]{0.120\linewidth}
        \caption{colobus}
    \end{subfigure}
    \begin{subfigure}[b]{0.120\linewidth}
        \caption{dwelling}
    \end{subfigure}
    \begin{subfigure}[b]{0.120\linewidth}
        \caption{lifeboat}
    \end{subfigure}
    \begin{subfigure}[b]{0.120\linewidth}
        \caption{quilt}
    \end{subfigure}
    \begin{subfigure}[b]{0.120\linewidth}
        \caption{trombone}
    \end{subfigure}

    \vspace{-10pt}
    \caption{
        Examples of ``easy'' and ``hard'' samples, each column belongs to the same category.
        Top row: samples with less computation. Bottom row: samples with more computation.}
    \label{fig:easy_hard_samples}
\end{figure*}

\textbf{The impact of $N$ and $r$.}
We adopt different values of $N$ and $r$ to explore
their impacts on the performance. As shown in Tab.~\ref{tab:resnet result}, we set $N=2,3,4$,
while keep $r=0.7$ on DMNN-50. The model with $N=4$ obtains the lowest test error rate,
indicating that bigger $N$ can lead to more path selection choices and consequently better performance.
We further keep $N=2$ and change $r$ to 0.4, 0.5 and 0.6 respectively. Larger $r$ leads to more computational
cost that verifies the effectiveness of our resource-constrained mechanism.
The model with larger FLOPs rate gains higher performance since more computation units are involved.
The DMNN can achieve a better accuracy-efficiency trade-off in terms of the computational budgets.
We have not conducted more experiments on larger $N$ due to resource limitation. But we will explore the characteristic
of DMNN with larger $N$ in the future work.

\subsection{Visualization}
\begin{figure}
    \centering
    \includegraphics[height=3.0in]{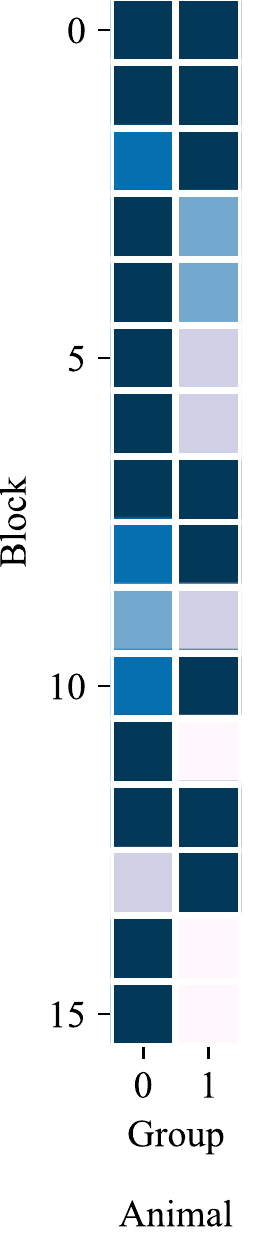}
    \hspace{10pt}
    \includegraphics[height=3.0in]{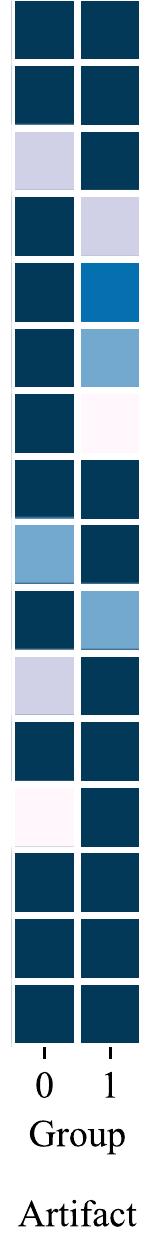}
    \hspace{10pt}
    \includegraphics[height=3.0in]{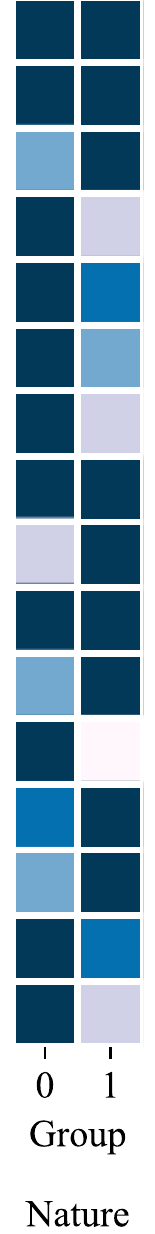}
    \hspace{10pt}
    \includegraphics[height=3.0in]{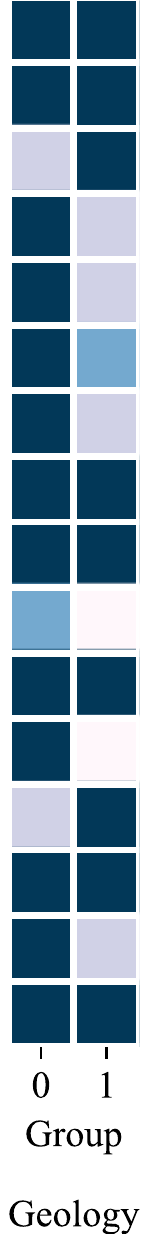}
    \hspace{10pt}
    \includegraphics[height=3.0in]{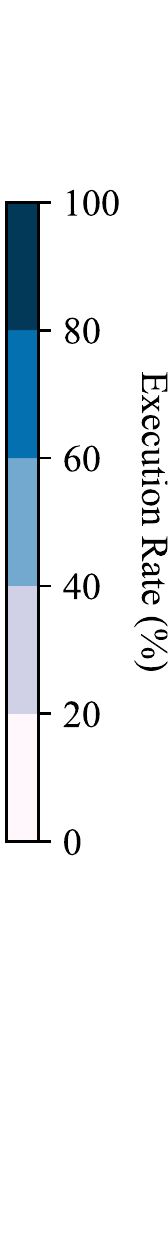}
    \vspace{-5pt}
    \caption{Execution rates of sub-blocks for different categories on DMNN-50 with $N=2$, $r=0.7$.}
    \label{fig:layer_rate}
    \vspace{-10pt}
\end{figure}

\textbf{Visualization of dynamic inference paths.}
The inference paths vary across images, which leads to different computation cost. Fig.~\ref{fig:dist}
shows the distribution of FLOPs on the ImageNet validation set using our DMNN-50 model with $N=2$, $r=0.7$.
The proportion of images with FLOPs in the middle is the highest, and images do occupy different computing
resources guided by computational constraint. We further visualize the execution rates of each
sub-block within the categories of animals, artifacts, natural objects, geological formations as shown
in Fig.~\ref{fig:layer_rate}. We can see that some sub-blocks, especially at the first two blocks of the
network, are executed all the time and the execution rates of other sub-blocks vary from categories.
One reason could be that different categories share the same shallow layers' features which are important
for classification. As the layer goes deeper, the semantic information of the features becomes stronger,
which depends on categories. 

\textbf{Visualization of ``easy'' and ``hard'' samples.}
We find that even samples of the same category would have different inference paths.
A reasonable explanation is that hard samples need more computation than easy ones.
Fig~\ref{fig:easy_hard_samples} shows examples of easy and hard samples with different actual FLOPs.
Although for some classes such as malamute and lifeboat, the ``hard'' samples are difficult than ``easy'' ones, for most classes, 
the quality gap is not indeed noticeable.  
We infer that it is because the definition of easy and hard samples mainly depends on the representation property of the
neural networks, 
rather than on the intuition of human beings. 

\section{Conclusion}
In this paper, we present a novel dynamic inference method called Dynamic Multi-path Neural Network (DMNN).
The proposed method splits the original block into multiple sub-blocks, making the network become more flexibility to
handle different samples adaptively.
We also carefully design the structure of the gate controller to get reasonable inference path, and introduce
resource-constrained lose to make full use of the representation capacity of sub-blocks.
Experimental results demonstrate the superiority of our method. 

%


{\small
\bibliographystyle{ieee}
\bibliography{egbib}
}

\end{document}